%% file: main.tex
\documentclass[sigconf]{acmart}
\setcopyright{none}
\renewcommand\footnotetextcopyrightpermission[1]{}

\usepackage{booktabs} 
\usepackage{multirow}
\usepackage{tabularx}
\usepackage{subfigure}

\begin{document}
\title{A Multi-Modal Approach to Infer Image Affect}

\author{Ashok Sundaresan, Sugumar Murugesan}



\author{Sean Davis, Karthik Kappaganthu, ZhongYi Jin, Divya Jain}

\author{Anurag Maunder}
\affiliation{%
  \institution{Johnson Controls - Innovation Garage}
  \city{Santa Clara} 
  \state{CA}
}

\renewcommand{\shortauthors}{JCI Garage}

\graphicspath{{./Figures/}}

\begin{abstract}
The group affect or emotion in an image of people can be inferred by extracting features about both the people in the picture and the overall makeup of the scene. The state-of-the-art on this problem investigates a combination of facial features, scene extraction and even audio tonality. This paper combines three additional modalities, namely, human pose, text-based tagging and CNN extracted features / predictions. To the best of our knowledge, this is the first time all of the modalities were extracted using deep neural networks. We evaluate the performance of our approach against baselines and identify insights throughout this paper.

\end{abstract}

%
%



\maketitle

\input{Introduction}
\input{Proposed_Approach}

\input{Results}
\input{Conclusions}
\bibliographystyle{ACM-Reference-Format}
\bibliography{bibliography} 

\end{document}

%% file: Introduction.tex
\section{Introduction}

Using machine learning to infer the emotion of a group of people in an image or video has the potential to enhance our understanding of social interactions and improve civil life.
Using modern computer vision techniques, image retrieval can be made more intuitive, doctors will be able to better diagnose psychological disorders and security will be able to better respond to social unrest before it escalates to violence.
Currently, even highly skilled professionals have a difficult time recognizing and categorizing the exact emotional state of a group.
Consistent, automatic categorization of the affect of an image can only be achieved with the latest advances in machine learning coupled with precise feature extraction and fusion.

Inferring the overall emotion that an image elicits from a viewer requires an understanding of details from vastly differing scales in the image.
Traditional approaches to this problem have focused on small-scale features, namely investigating the emotion of individual faces \cite{Goodfellow15, Reddy16, Yu15}.
However, both large-scale features \cite{Dhall15, Li16} and additional action/scene recognition must be represented in the description of the image emotion \cite{Mathews16, Shin16}.
Each of these feature extraction methods, or modalities, cannot capture the full emotion of an image by itself.

Previous research has investigated a combination of facial features, scene extraction \cite{Li16} and even audio tonality \cite{Sun16}.
This paper combines three additional modalities with commonly used facial and scene modalities, namely,  human pose, text-based tagging and CNN extracted features / predictions.
To the best of our knowledge, this is the first time all of the modalities were extracted using deep neural networks with the exception of the baseline CENsus TRansform hISTogram (CENTRIST) \cite{Wu11} model.

Our approach combines the top-down and bottom-up features. Due to the presence of multiple subjects in an image, a key problem that needs to be resolved is to combine individual human emotions to a group level emotion. The next step is to combine multiple group level modalities. We address the first problem by using a Bag-of-Visual-Words (BOV) based approach. Our BOV based approach comprises developing a code book using two well known clustering algorithms: k-means clustering and a  Gaussian mixture model (GMM) based clustering. The code books from these  clustering algorithms are used to fuse individual subjects' features to a group image level features 

We evaluate the performance of individual features using 4 different classification algorithms, namely, random forests, extra trees, gradient boosted trees and SVM. Additionally, the predictions from these classifiers are used to create an ensemble of classifiers to obtain the final classification results. While developing the ensemble model, we also employ predictions from a individually trained convolutional neural network (CNN) model using the group level images. 

%% file: Proposed_Approach.tex
\section{Modalities}
\input{Chapters/Modalities}
\input{Chapters/Feature_Fusion}

%% file: Chapters/Modalities.tex
\input{Chapters/Face_Features}
\input{Chapters/Centrist}
\input{Chapters/Human_Pose}
\input{Chapters/Image_to_Tag}
\input{Chapters/CNN}

%% file: Chapters/Face_Features.tex
\subsection{Facial Feature Extraction}
Facial features are extracted individually from isolated human images on an image-by-image basis.
First,  a Faster R-CNN \cite{Ren15} is employed to extract the each of the full human frames from the image.
These extracted frames are used for both the pose estimation and the facial feature extraction.
Each face is then extracted and aligned with the frontal view using Deepface \cite{Taigman}.
Once the faces are isolated and pre-processed, a deep residual network (ResNet) \cite{He15} is employed to extract facial features.

\subsubsection{Human Frame Extraction}
Isolation of each significant human frame is performed on the image using a Faster R-CNN \cite{Ren15}.
The model is trained on the Pascal VOC 2007 data set \cite{Voc07} using the very deep VGG-16 model \cite{Simonyan14}.
The procedure for building and training this model is presented in the original Faster R-CNN manuscript \cite{Ren15} and will not be detailed here.

\subsubsection{Deepface}
For extracting and aligning human faces in the images, Facebook's DeepFace \cite{Taigman} algorithm is utilized.
The alignment is performed using explicit 3D facial modeling coupled with a 9-layer deep neural network.
For more details on the implementation and training of the model, please see \cite{Taigman}.

\subsubsection{ResNet based feature extraction}
The ResNet model is trained with the Face Emotion Recognition (FER13) dataset \cite{Goodfellow15}, which contains 35,887 aligned face images with a predefined training and validation set.
Each face in FER13 is labeled with one of the following 7 emotions: anger, disgust, fear, happiness, sadness, surprise or neutral.
To balance training time and efficacy, the ResNet-50 topology from \citenum{He15} is implemented.
The ResNet model is trained from scratch, with an initial learning rate of 0.1, a learning rate decay factor of 0.9 which decays every 2 epochs and a batch size of 64.
This network architecture is able to achieve a 62.8\% top-1 accuracy on the validation FER13 set when predicting the emotion from an image.
In lieu of a fully connected layer, ResNet-50 uses global average pooling as its penultimate layer.
The feature set, which is a vector of 2048 elements, is the output from the global average pooling layer when running inference on the isolated faces from each image.

%% file: Chapters/Centrist.tex
\subsection{Centrist}

\begin{figure*}[htp]
  \centering
  \subfigure[Image from emotiw2017 training dataset]{\includegraphics[scale=1.1]{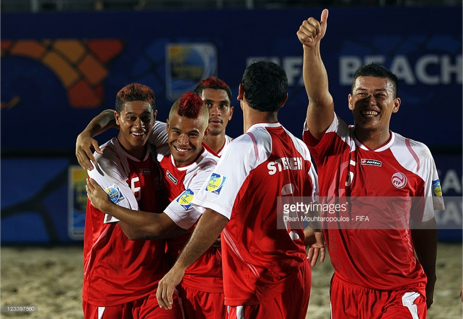}}\quad
  \subfigure[Centroid Transform of image to the left]{\includegraphics[scale=1.1]{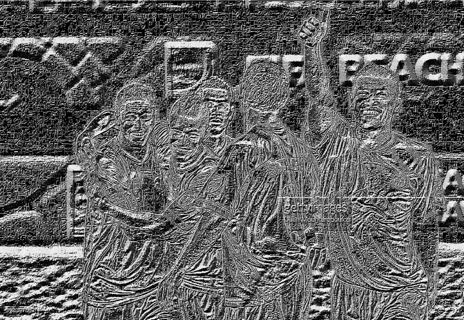}}
  \caption{Centrist illustration}
  \label{fig:soccer}
\end{figure*}

The CENsus TRansform hISTogram (CENTRIST) \cite{Wu11} is a visual descriptor predominantly used to describe topological places and scenes in an image \cite{6825882, 6696653, 7298721, 7350829}. Census Transform, in essence, is calculated by comparing a pixel's gray-scale intensity with that of its eight neighboring pixels, encoding the results as 8-bit binary codeword, and converting the codeword to decimal (Fig. \ref{fig:soccer}). CENTRIST, in turn, is the histogram of the Census Transform calculated over various rectangular sections of the image via Spatial Pyramid techniques \cite{1641019}. By construction, CENTRIST is expected to capture the high-level, global features of an image such as the background, the relative locations of the persons, etc. CENTRIST is adopted by the challenge organizers as the baseline algorithm for emotion detection. The baseline accuracy provided by the organizers is 52.97$\%$ on the Validation set and 53.62$\%$ on the Test set. To achieve these accuracies, a support vector regression model was trained using features extracted by CENTRIST. In our work, we use CENTRIST as a scene-level descriptor and build various modalities on top of it to extract complementary features from the image.


%% file: Chapters/Human_Pose.tex
\subsection{Human Pose}
The intention of including pose modality in this emotion detection task is to detect similar and dissimilar poses of the individuals in the image and capture any effect of these poses may have towards the group emotion. The pose features are expected to work only as an indirect complement to features from other modalities. Human pose estimation is a regression problem in which the location of specific key points in a human body are predicted. In literature several neural network and non-neural network methods have been used. State of the art results on PASCAL VOC have been demonstrated by Gkioxari et.al. \cite{gkioxari2014using}, \cite{poseactionrcnn}. For this work, we utilize the work presented in \cite{poseactionrcnn} which is based on a R-CNN. This method obtained a mean AP of 15.2\% PASCAl VOC 2009 validation dataset. The method uses AlexNet \cite{krizhevsky2012imagenet} and builds an R-CNN \cite{girshick2014rich} with region proposals generated by \cite{arbelaez2014multiscale} and fine tune from ImageNet pertained model on 1000 classes. The R-CNN in this work is trained to predict key points to be within a distance of .2 H (H is the height of the torso) from the ground truth. During test time, the fc7 embeddings are used as features to combine with other modalities.

%% file: Chapters/Image_to_Tag.tex
\subsection{Image Captions}
\subsubsection{ClarifAI}
One of the modalities we used in our model is to generate captions for images and use captions to infer emotion. As a first step towards that, we first proceeded to use commercially available ClarifAI API \cite{clarifai} to answer the following preliminary questions: 

\begin{itemize}
    \item Whether reasonably accurate captions could be generated for group-level emotion images
    \item Whether captions are descriptive of underlying emotions
\end{itemize}

\begin{figure*}[htp]
  \centering
  \subfigure[Traditional: 0.99906313, Festival: 0.9946226, Culture: 0.99122095, Religion: 0.99001175]{\includegraphics[scale=1.05]{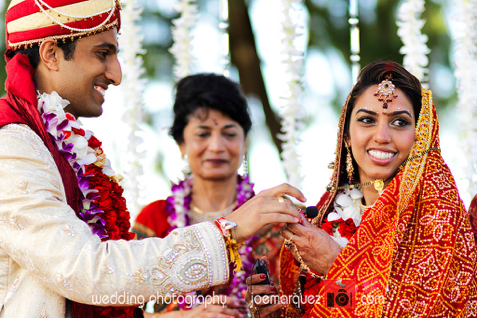}}\quad
  \subfigure[People: 0.97925866, Drag Race: 0.964833, Police:0.9458773, Battle:0.9399004]{\includegraphics[scale=1.1]{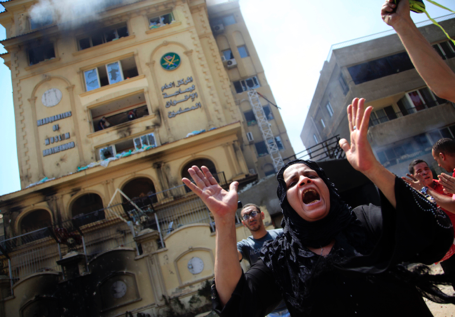}}
  \caption{Top four tags generated from ClarifAI in `tag:probability' format. 
  Image source: emotiw2017 training dataset}
  \label{fig:clarifai_fig}
\end{figure*}

\begin{figure}[h!]
\includegraphics[scale=0.5]{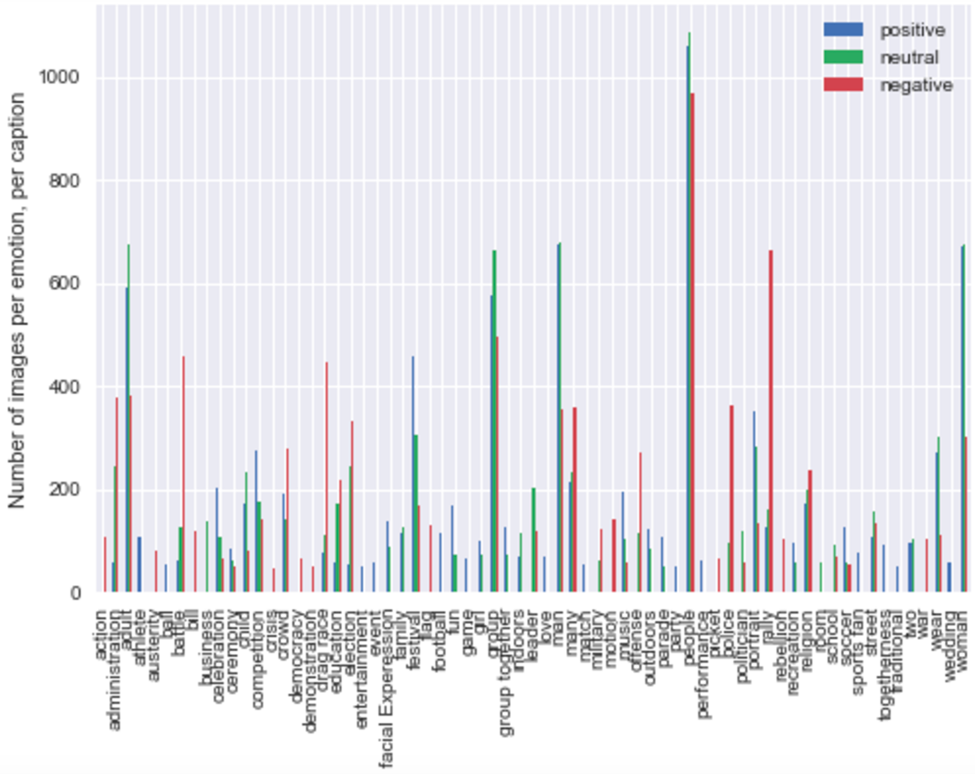}
\caption{Distribution of top tags generated by Clarifai, across three emotion classes for images in emotiw 2017 training dataset}
\label{fig:clarifai_tags}
\end{figure}

\noindent Fig. \ref{fig:clarifai_fig} illustrates the top four tags generated by Clarifai for some of the images in the Emotiw2017 Training Dataset. These results along with those generated for several other images in the dataset provide us confidence that a deep learning model can be reasonably accurate in generating tags for images, thus answering the first question above. As for the second question, we refer to Fig. \ref{fig:clarifai_tags}. Here, the distribution of top tags generated by Clarifai across three emotion classes is plotted. Note that, for several tags such as `administration', `election', `competition', there is a notable difference in prior probability of occurrence given each class. This could translate to discriminative probability of classes given these tags, assuming uniform prior on the classes. This observation is even more pronounced for more obvious tags such as `festival' and `battle'. This answers the second question above, and motivates us to explore further.

\subsubsection{im2txt}

Having motivated ourselves that image tags can be of value in inferring emotions on a group-level image, we turned our attention to the more general `image to caption' models. Towards this, we focused on the {\it im2txt} TensorFlow open-source model \cite{cshallue}. This model combined techniques from both computer vision and natural language processing to form a complete image captioning algorithm. More specifically, in {\it machine translation} between languages, a Recurrent Neural Network (RNN) is used to transform a sentence in `source language' into a vector representation \cite{cho, bahdanau, sutskever}. This vector is further fed into a second RNN that generates a target sentence in the `target language'. In \cite{7505636}, the authors derived from this approach and replaced the first RNN with a visual representation of an image using a deep Convolutional Neural Network (CNN) \cite{sermanet} until the pen-ultimate layer. This CNN was originally trained to detect objects, and the pen-ultimate layer is used to feed the second RNN designed to produce phrases in the original machine translation model. This end-to-end system is further trained directly on images and their captions to maximize the likelihood that the description on an image best matches the training descriptions for that image. The open-source model im2txt \cite{cshallue} is an implementation of the algorithm in \cite{7505636}. We specifically used a Dockerized version of the im2txt model available in \cite{siavash}.

\begin{figure*}[htp]
  \centering
  \subfigure[a woman sitting on a couch with a child in front of her, a woman sitting on a couch with a child on her lap, a woman sitting on a couch with a child in her lap]{\includegraphics[scale=0.46]{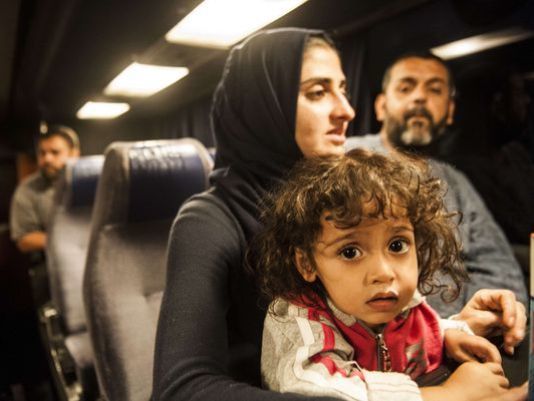}}\quad
  \subfigure[a group of people sitting around a table, a group of people sitting around a table with wine glasses, a group of people sitting around a table with food]{\includegraphics[scale=0.41]{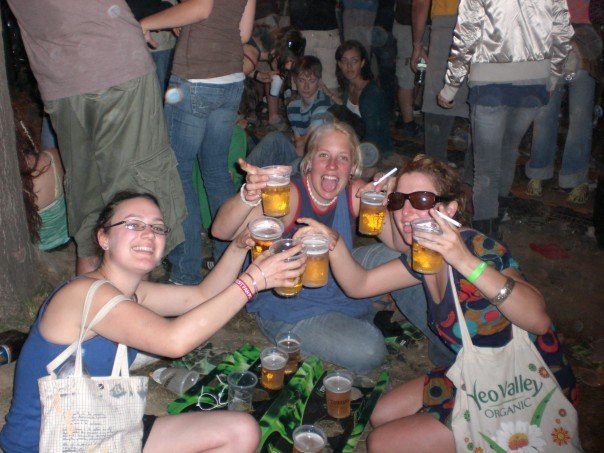}}
  \caption{Captions generated by im2txt model. Image source: emotiw2017 training dataset}
  \label{fig:im2txt}
\end{figure*}

Fig. \ref{fig:im2txt} illustrates the captions generated by im2txt in \cite{siavash} for some of the images in the training dataset. In our approach, these captions are further encoded into sparse bag-of-word vectors \cite{Zhang2010}. For instance, if the dictionary passed to the im2txt model is made of words $\{w_1, w_2, \ldots w_n\}$. Then if, for an image, caption is $w_3 w_4 w_2 w_4$, the bag-of-words representation of the image is: $\{w_1:0, w_2:1, w_3:1, w_4:2, w_5:0 \ldots w_n:0\}$. This vector is further normalized during the concatenation stage, effectively converting this into a term frequency representation \cite{CHOW200912023}.

%% file: Chapters/CNN.tex
\subsection{CNN}
CNNs are a proven technique for image classification. The complexity and sparsity of affection related features in images suggest a CNN only solution will need a deeper and wider network and a large training dataset. However, even with a relatively small training dataset, CNNs can still be used as an effective modality for feature extraction. To reduce overfitting, the Resenet18 architecture \cite{He15} is employed as the CNN and two models are built with the training dataset: one is trained on original colored image and one is trained on grayscale images converted from the original ones. In cross-validation, the color-model has an accuracy of 52\% and the gray-model 50\%. This suggests color contains affection information. For the remaining of the paper, we will only refer the color model as our CNN model.

%% file: Chapters/Feature_Fusion.tex
\section{Feature Aggregation}

As previously discussed in Section ~\ref{sec:sec3}, given a group level image, our methodology consists of extracting scene related features using CENTRIST, facial features using RESNET, human pose features using a pre-trained NN and BOW descriptors using the im2txt neural network. In this section, we describe our strategy for combining features extracted from all the group level images to build the training data for classification. 

Our feature combination strategy involves concatenating all the features for each group level image. It is important to note that, in this problem of group level emotion recognition, feature concatenation is not straight-forward. The feature vectors extracted from CENTRIST and the bag-of-words extracted from the im2txt NN are already on a per image basis.However, this is not the case for the facial and pose features since these features are extracted for each human in the images and need to be aggregated across all the humans for each group level image, before concatenation.

Inspired by [references], we adopt a bag-of-visual-words (BOV) based approach to construct image level aggregated feature vectors using the facial and pose features for each group level image evaluated using the methods described in Section and Section. The BOV based feature transformation is carried out separately for both facial and pose features. 

In our BOV based approach, the feature vector corresponding to each face or pose, is regarded a visual word as a result of which the number of visual words is equal to the total number of humans extracted over all the group level images. Having defined the visual word, the BOV based feature aggregation is described as follows.
\begin{enumerate}
\item \underline{k-means Based Methodology}
The set of visual words is clustered using the k-means algorithm to reduce the vocabulary which consists of the cluster labels, also referred to as visual codes. From clusters resulting from the k-means clustering algorithm, we develop three group level feature vectors as follows. 

\begin{enumerate}
\item	A term frequency (TF) matrix which consists of the associating each face in the group level image to a visual code and the counting the occurrence of each visual code in the vocabulary. The values are also normalized by dividing each count by the total number of faces (visual words) in the image. Denoting the raw count of a visual code $l$ for a given image $m$ as $t_{l,m}$, the TF is simply given as,
\begin{equation}
tf(i,m) = \frac{ t_{l,m}} {\sum_{l^{\prime} \in m} t_{l^ {\prime},m}}
\end{equation}

\item The second feature, that we evaluate using the cluster center is the vector of locally aggregated descriptors (VLAD). VLAD is a feature encoding method that takes into account the strength of association of each visual word to its cluster center. 

Defining the strength of association of each visual word ${w_i}$ ($i=1,\ldots,n$) to cluster $k$ as $q_{ik}$, where $q_{ik} \geq 0$ and $\sum_{i=1}^{n} q_{ik} = 1$, the VLAD encoding using the cluster centroids $\mu_{k}$ is defined as,

\begin{equation}
v_k = \sum_{i=1}^{n} q_{ik} (w_{ik} - \mu_k)
\end{equation}

The primary advantage of VLAD over the TF matrix is that more discriminative property is added in our feature vector by taking the difference of each descriptor from the mean in its voronoi cell. This first order statistic adds more information in the feature vector and may give us better discrimination during classification. The VLAD encodings are usually normalized (e.g. $l^2$ normalization) before usage.
	
\item The third feature vector, resulting from the k-means clustering algorithm is a weighted average (WA) of the cluster centers for each visual word in the image. The  weights are nothing but the normalized term frequencies corresponding to each visual word.
\end{enumerate}

\item \underline{GMM Based Methodology}  In addition to the k-means clustering method, we also perform a dimensionality reduction of the visual codes using GMM. The aggregation  involves computing the posterior probability of each component for a given visual word, which is also defined as the responsibility that a particular component takes in explaining that visual word. The responsibilities of a given component resulting from all the faces in a given image are averaged to compute the group level image responsibility for that component. Performing the same computation for all the components using the visual words in a given image results in an image level aggregated feature vector.

\end{enumerate}

It can be noted that all the four feature vectors described above, namely, the TF matrix, VLAD, the weighted average of the cluster centers and the GMM based features have dimensions $N \times K$, where $K$ is either the number of clusters for the k-means algorithm or the number of Gaussians for the GMM model. Since all the features are now consistent in terms of their dimensions and are group-image level features, they can be easily concatenated. The series of above described steps are carried out for both the training and the validation image sets to construct the features for classification.

%% file: Results.tex
\section{Classification Results}
\label{sec:Classification}


Given the feature vectors for group level images corresponding to both the training and the validation set, the next task is to build a classification model which can assign a given image to one of the three classes: positive, negative or neutral. To build the classification model, we adopt a two tier approach: tier-1, consisting of learning multiple independent classifiers and, tier-2 learning from tier-1 classifiers. Such an ensemble approach has previously been used in many data science competitions.

The first tier involves training individual classifiers on each of the different modalities and optimizing their parameters that result in the maximum validation accuracy. To this end, four different well-known classifiers are considered, namely, a random forest (RF) classifier, gradient boosted tree (GBT) based classifier, support vector machine (SVM) classifier and a variant of random forests called as extra trees (ET).  Each of the classifiers considered are inherently different in the sense that they may learn different characteristics of the feature set to perform classification. For example, the RF classifier may be well suited for a particular group level emotion whereas the performance of other classifiers may be below par. In a different scenario, we may have a situation where the SVM based classifier is better suited. Motivated by this possibility, our second tier model consists of building an ensemble of the above mentioned classifiers using a technique referred to as stacking. Also, note that the input features for each of these classifiers are also made up of different modalities thereby providing a diversity of information for the second tier algorithm to learn. A Logistic regression model is used as the second tier classifier.

In this section, we provide more details about the features extracted from the group level images using the methods outlined in the previous sections. 

From the training dataset of 3630 images, the human detection and face detection pipeline extracted 19464 faces. For the validation set we extracted 11696 faces from 2066 images. For these images, we use vocabulary sizes of 200, 500 and 1000 to compute the BOV derived features. The performance of the final classification was best for a vocabulary size of 1000 and hence for conciseness we will explain the rest of the results based on this vocabulary size.  The list of all feature vectors used in tier-1, along with their dimensions are shown in Table~\ref{tab:features_table}. 

\begin{center}
\begin{table}
\begin{tabular}{|c|c|c|}
\hline
Modalities & Feature & Dimension \\
\hline
\multirow{4}{4em}{Face emotion} & Term Frequency & 1000\\
& VLAD & 1000\\
&Weighted Average of Cluster Center & 2048\\
&GMM & 512\\
\hline
\multirow{4}{4em}{Pose} & Term Frequency & 1000\\
& VLAD & 1000\\
&Weighted Average of Cluster Centers & 2048\\
&GMM & 512\\
\hline
{Scene} & Centrist & 7905\\
\hline
{Image to Text}& Bag of Words & 364\\
\hline
\end{tabular}
\caption{Features extracted for tier-1 classifiers}
\label{tab:features_table}
\end{table}
\end{center}


Next, we determine the performance of each of the modalities separately with different classifiers. These results are listed in Tables~\ref{tab:centrist_performance} to  \ref{tab:pose_performance}.
\begin{center}
\begin{table}
\begin{tabular}{|c|c|c|c|c|}
\hline
   & RF & ET & GBT & SVM \\
\hline
Training & .99 & .98 & .98 & .95 \\
Validation & .55 &.53 & .52 & .50 \\
\hline
\end{tabular}
\caption{Centrist Based Features Classification Performance}
\label{tab:centrist_performance}
\end{table}
\end{center}

\begin{center}
\begin{table}
\begin{tabular}{|c|c|c|c|c|c|}
\hline
   & & RF & ET & GBT & SVM\\
\hline
TF & Training & .99 & .99 & .90 & .95 \\
& Validation & .56 &.58 & .57 & .56 \\
\hline
VLAD & Training & .99 & .99 & .99 & .97 \\
& Validation & .58 &.58 & .59 & .53 \\
\hline
Weighted Average & Training & .99 & .99 & .99 & .99 \\
& Validation & .60 &.58 & .57 & .58 \\
\hline
GMM & Training & .99 & .99 & .90 & .87 \\
& Validation & .56 &.56 & .55 & .58 \\
\hline

\end{tabular}
\caption{Face Emotion Based Features Classification Performance}
\label{tab:face_emotion_performance}
\end{table}
\end{center}

\begin{center}
\begin{table}
\begin{tabular}{|c|c|c|c|c|c|}
\hline
   & & RF & ET & GBT & SVM\\
\hline
TF & Training & .99 & .99 & .99 & .98 \\
& Validation & .39 &.40 & .42 & .38 \\
\hline
VLAD & Training & .99 & .99 & .99 & .99 \\
& Validation & .41 &.40 & .41 & .40 \\
\hline
Weighted Average & Training & .99 & .99 & .99 & .99 \\
& Validation & .40 &.41 & .41 & .42 \\
\hline
GMM & Training & .99 & .99 & .95 & .96 \\
& Validation & .40 &.43 & .41 & .40 \\
\hline
\end{tabular}
\caption{Pose Based Features Classification Performance}
\label{tab:pose_performance}
\end{table}
\end{center}

The classification performance for the features apart from pose meet or exceed the benchmark performance \cite{Dhall15}.  Since the pose based features is poor, we do not include it in the concatenated features. Also, it is interesting to note that the performance of the concatenated features is better than the performance of individual features.  Among the individual modality features, the weighted average features for face emotion modality using random forests achieves the best validation accuracy of 60.22\%.  We observed that the concatenated features provide the best classification accuracy on the validation set (64\%). Hence, we use it as the basis for the stacking classifier.

The probabilities predicted from the individual classifiers with the concatenated feature set as the input, was used to train a three class Logistic Regression classifier to perform stacking. This classifier improves the validation accuracy by 4\% resulting in an overall validation accuracy of 68\%.

%% file: Conclusions.tex
\section{Conclusions and Future Directions}
The problem of determining the affect of a group of people is considered in this paper. A multi-modal approach consisting of features extracted from the scene, human face, human pose and image tags is adopted. To extract the human face based features, we developed a pipeline that consists of R-CNNs for extracting humans from the group level images, a CNN \cite{Taigman} to crop the faces and a conventional face alignment method that uses regression trees.

One of the main contributions of this work is the training and usage of deep neural networks (DNNs) to extract face and pose features. Furthermore, the DNNs were trained on external datasets that contained more number of images suitable for learning the large number of parameters in a DNN. This approach overcomes of the inadequacy of the current dataset, which does not have sufficient samples to train a DNN. Additionally, we also employed the bag-of-visual-words based approach to translate multiple human level features to group level features. Using a combination of these group level features from multiple modalities, a validation accuracy of  64\% percent was achieved. Finally, a stacking methodology was employed to build an ensemble of classifiers which resulted in a validation accuracy of 68\%. 

One key observation that was noted in this work is that our methodology was not able to exploit pose based features in an optimal fashion. This needs to be investigated further. Also, improvements could be made to the methodology adopted for the image to text based feature extraction. our current method extracted too few image tags which could be limiting the performance of the classifier.

Weak supervision \cite{Ratner17} has been proven as an effective way of obtaining additional labeled data. As for future work, we are planning to use google reverse-image search as our first labeling function to collect similar images as the one in the training dataset. In the same process, context information such as human annotated tags and summary information will also be gathered for each newly collected image. We will then apply a second NLP based labeling function to label the new images. Our initial experiment shows that with the newly labeled images, the Restnet18 model has 1\% improvement in terms of accuracy.